\documentclass[fleqn,10pt]{template_style/wlscirep}
\usepackage{amsmath,amsfonts,amssymb}
\usepackage[utf8]{inputenc}
\usepackage[T1]{fontenc}
\usepackage{enumitem,float,lineno,hhline}
\usepackage{color,hyperref,multirow}
\usepackage{framed,algorithmic,algorithm}

\usepackage{siunitx}
\newcommand{\degr}{{^\circ}}

\title{Parallel-beam X-ray CT datasets of apples with internal defects and label balancing for machine learning}

\author[1,$\dag$,*]{Sophia Bethany Coban}
\author[1,$\dag$,*]{Vladyslav Andriiashen}
\author[1,2,$\dag$,*]{Poulami Somanya Ganguly}
\author[1,3]{Maureen van Eijnatten}
\author[4]{Kees Joost Batenburg}
\affil[1]{Centrum Wiskunde \& Informatica, Science Park 123, 1098 XG Amsterdam, The Netherlands.}
\affil[2]{Mathematical Institute, Leiden University, Niels Bohrweg 1, 2333 CA Leiden, The Netherlands.}
\affil[3]{Eindhoven University of Technology, Groene Loper 3, 5612 AE Eindhoven, The Netherlands.}
\affil[4]{Leiden Institute of Advanced Computer Science, Niels Bohrweg 1, 2333 CA Leiden, The Netherlands.}
\affil[*]{\small corresponding author(s): S.B. Coban (s.b.coban@cwi.nl), V. Andriiashen (vladyslav.andriiashen@cwi.nl), P.S. Ganguly (poulami.ganguly@cwi.nl@cwi.nl).}
\affil[$\dag$]{\small These authors contributed equally to this work.}

\begin{abstract}
We present three parallel-beam tomographic datasets of 94 apples with internal defects along with defect label files. The datasets are prepared for development and testing of data-driven, learning-based image reconstruction, segmentation and post-processing methods. The three versions are a noiseless simulation; simulation with added Gaussian noise, and with scattering noise. The datasets are based on real 3D X-ray CT data and their subsequent volume reconstructions. The ground truth images, based on the volume reconstructions, are also available through this project. Apples contain various defects, which naturally introduce a label bias. We tackle this by formulating the bias as an optimization problem. In addition, we demonstrate solving this problem with two methods: a simple heuristic algorithm and through mixed integer quadratic programming. This ensures the datasets can be split into test, training or validation subsets with the label bias eliminated. Therefore the datasets can be used for image reconstruction, segmentation, automatic defect detection, and testing the effects of (as well as applying new methodologies for removing) label bias in machine learning. 
\end{abstract}
\begin{document}

\flushbottom
\maketitle

\thispagestyle{empty}

\section*{Background \& Summary}
X-ray computed tomography (CT) is a versatile non-destructive imaging technique with a wide range of applications in clinical~\cite{liguori2015emerging}, industrial~\cite{DECHIFFRE2014655} and scientific settings~\cite{mees2003applications, dierolf2010ptychographic, egan20153d, morigi2010application}. The basic working principle of CT is based on the acquisition of raw measurement data from different angles, via which an image can be reconstructed by solving the resulting inverse problem. Various analytical and iterative methods have been proposed for this task. However, when the acquired measurement data is limited, e.g. due to reduced X-ray flux (low-dose), few angles or a limited angular range, analytical methods result in inaccurate and noisy reconstructions, while the computational load of more sophisticated iterative methods is often too demanding for high-throughput settings.

In such cases, data-driven methodologies such as deep neural networks have shown promising results in tackling the shortcomings of traditional analytical and iterative reconstruction methods~\cite{bubba2019learning, adler2018learned, arridge2019solving}. However, certain challenges remain, such as the large amounts of labelled data needed to develop and train deep neural networks. Particularly the absence of segmentation labels based on accurate ground truth images is a major limitation in practical settings, as such labels typically needs to be obtained manually. Furthermore, it remains unclear which deep learning-based method results in the best reconstruction accuracy but also generalizes well to new, unseen samples. This means that the method should not over-fit on the training data, with the risk of significant under-performance when transferring the method to different types of imaging data and applications\cite{reitermanova2010data}. Since it can be difficult or even impossible to avoid all types of bias during data acquisition~\cite{biondetti2020name}, techniques need to be developed to mitigate the impact of bias prior to or even during training of deep learning methods.

A common factor in the successful applications of deep learning is the abundance of labelled training data~\cite{deng2009imagenet}. To enable the integration of deep learning in image reconstruction methods, it is therefore essential to make large tomographic dataset -- including the ground truth reconstructions -- publicly available to the scientific community. To date, a limited number of medical imaging datasets has been made available in the context of Grand Challenges. Examples of such challenges are the low-dose CT Challenge initiated by the Mayo clinic~\cite{mccollough2017low} and the fast MRI challenge~\cite{zbontar2018fastmri}.
However, in most of these challenges only dozens of images have been made available.  A larger medical tomographic dataset that has recently been made available is the LoDoPaB dataset~cite{leuschner2019lodopab}, which contains over 40000 scan slices from around 800 patients selected from the LIDC/IDRI Database with medical chest CT scans~\cite{armato2011lung}. In addition, a non-medical dataset consisting of 42 walnuts acquired using a cone-beam CT scanner has been published for high cone-angle artefact reduction \cite{walnut-dataset}. Nevertheless, these datasets remain limited to specific use cases and CT applications.

As a case that is representative for a broad range of CT applications, in which products are scanned in large numbers and labelled data are difficult to obtain, this work focuses on abnormality detection in fruit~\cite{van2020nondestructive}. We describe a dataset of 72192 two-dimensional CT images of apples with internal defects, and 50 corresponding parallel-beam projections for each image. Three different types of projection data are available: 1) noise-free projections; 2) projections with added Gaussian noise, and 3) projections with added scattering based on the apple volume for a subset of the images. The projections can be used to simulate various sparse or limited-view angular range tomographic problems, in which the volumetric apple reconstructions serve as ground truth. We show an example of such a tomographic problem, namely limited-view angular sampling, in Technical Validation. Due to the large number of available projections and corresponding ground truth reconstructions, our dataset is well-suited to train deep neural networks for CT reconstruction tasks.

As an additional contribution of this work, we introduce two fast and effective methods to eliminate label bias when splitting data for machine learning purposes. We describe an empirical splitting method and a splitting method through integer programming that can be used to split a labelled dataset into balanced training, validation and test subsets. In this context, balanced means that the internal defect percentages in the apples are equally distributed across the different splits, and therefore there is no bias towards a specific apple, and more importantly, a specific defect within the apple.

\section*{Methods}

\subsection*{Tomographic data acquisition and reconstruction of the apples}

The tomographic datasets of 94 apples were acquired using the custom built and highly flexible CT scanner at the FleX-ray Laboratory, developed by TESCAN-XRE NV, managed by the Computational Imaging group at Centrum Wiskunde \& Informatica, Amsterdam \cite{Flexray2020}. 

The scanner consists of a cone-beam microfocus X-ray point source that projects polychromatic X-rays onto a $1944\times 1536$ pixels, 14-bit, flat detector panel. The scanned apples were of similar sizes, thus the tube or geometry settings were determined for the shape and material density of the first apple, and remained unchanged for the rest of the scans. The tube voltage and output power were 45kV/45W, respectively. Each apple data was captured at sample resolution of 54.2$\mu$m (magnification: 1.4) over a 360$\degr$ in circular and continuous motion. A total of 1200 projections (or radiographs) were collected, which were distributed evenly over the full circle (the angular increment was 0.3$\degr$). Each projection was acquired with 100ms exposure time. In addition, an open-shutter (flat-field) and closed-shutter (dark-field) images were taken before each data collection in order to normalise the acquired data and therefore accurately capture the material attenuation distribution free of any tube- or detector-dependent artefacts. The total scanning time for each apple was just over 4 minutes. 

Prior to any ground truth reconstruction, each dataset was normalised such a way that satisfies the Beer-Lambert Law \cite{beer,lambert}, 
\begin{align}\label{eq:beerlambert}
    P = \int_L \mu \delta L = -\log\frac{I}{I_0},
\end{align}
where $P$ denotes the collected projections, which are the definite integral (or the sum) of material attenuation coefficient, $\mu$, over the sample thickness, $L$, (the "ray-sum"); $I_0$ the difference between each projection and the open-shutter images, $I$ the difference between closed-shutter and open-shutter images, and $\log$ the natural logarithm. The "ray-sum" in Equation~\ref{eq:beerlambert} is typically expressed using the Radon transform\cite{radon86}, where the image is recovered via a back-projection (inverse Radon transform) and an appropriate filter (often the ramp filter)\cite{FNattarer,FDK84}. The image can also be recovered via an iterative reconstruction method with an appropriate discretization of the geometry set-up. This allows the problem of image reconstruction to be expressed as a set of linear equations, $Ax = b$, where $x$ is the image to reconstruct, $b$ is the vectorized projection data, and $A$ is the geometry set-up obtained with a ray tracing algorithm\cite{jacobs_raytracing,siddon_raytracing}. 

Adjustments to any rotational mismatch of the sample (also known as centre-of-rotation correction) is achieved via the algorithm described in \cite{tliu_COR09}, applied to the geometry prior to each reconstruction. The implementation of this correction algorithm is included in \cite{beadscodes}. 

For the ground truth generation, we employed an iterative reconstruction method, namely, the conjugate gradient least squares (CGLS) \cite{YSaad_CGLS,CGLSBjorck}. The reconstructions were obtained using the CGLS algorithm with 15 iterations, employing the same set-up as described in \cite{beadswork,beadscodes}. Reconstructions were computed on a Linux machine with specifications 2.1GHz, 64-bit 16-core CPU (2x 8-core Intel Xeon Silver 4110) using a single GPU with 11GB RAM (NVIDIA Geforce GTX 1080Ti). Data processing and reconstruction for each apple took around 7 minutes to complete.

\subsection*{Ground truth generation}

\begin{figure}[ht]
\centering
\includegraphics[scale=0.4]{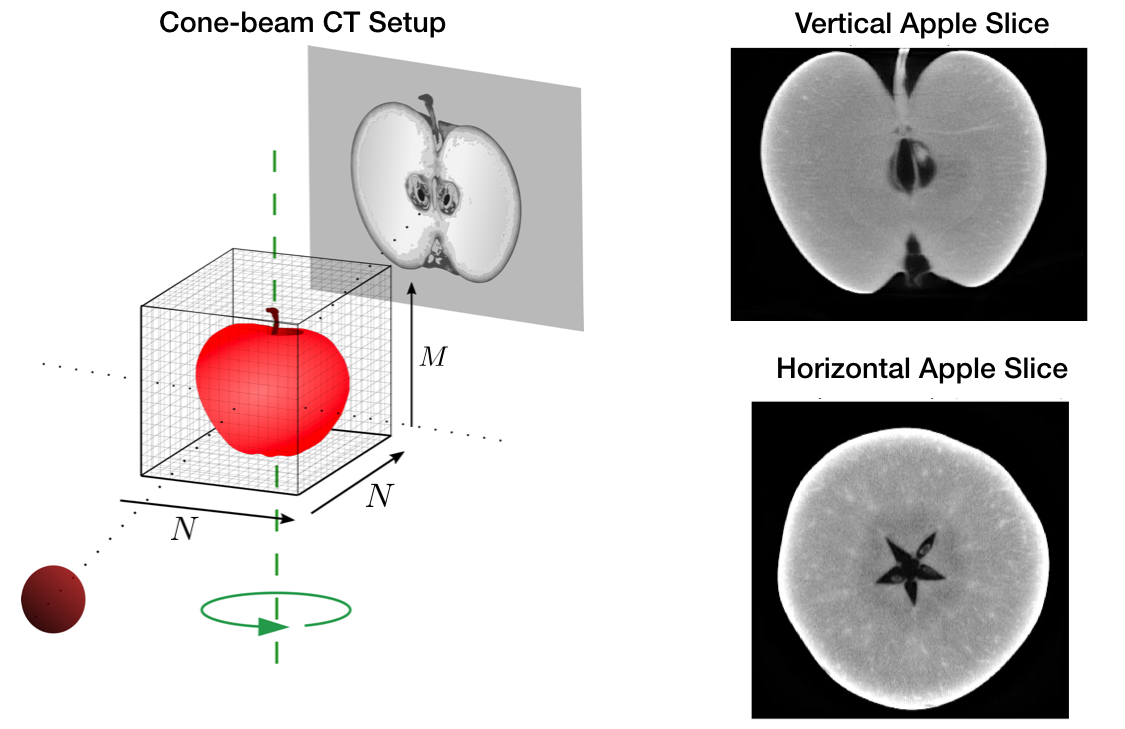}
\caption{An illustration of the circular cone-beam CT geometry set up is shown on the left; example slices for the vertical and horizontal directions in the reconstructed volumes on the right.} 
\label{fig:applerecon-info}
\end{figure}

The ground truth slices were generated based on the cone-beam volumetric CGLS reconstruction of each apple. We note here that when we mention an "apple slice", we refer to a horizontal 2D slice through the 3D reconstruction of an apple. Otherwise, we specify that it is a \textit{vertical} apple slice through the reconstruction. Figure~\ref{fig:applerecon-info} shows the cone-beam tomographic set-up in which the apples were scanned, and the orientation of the apple slices in vertical and horizontal dimensions. Note that the voxel grid in the figure is denoted by $N\times N\times M$, which in our case was $972\times 972\times 768$.

\begin{figure}[ht]
\centering
\includegraphics[scale=0.35]{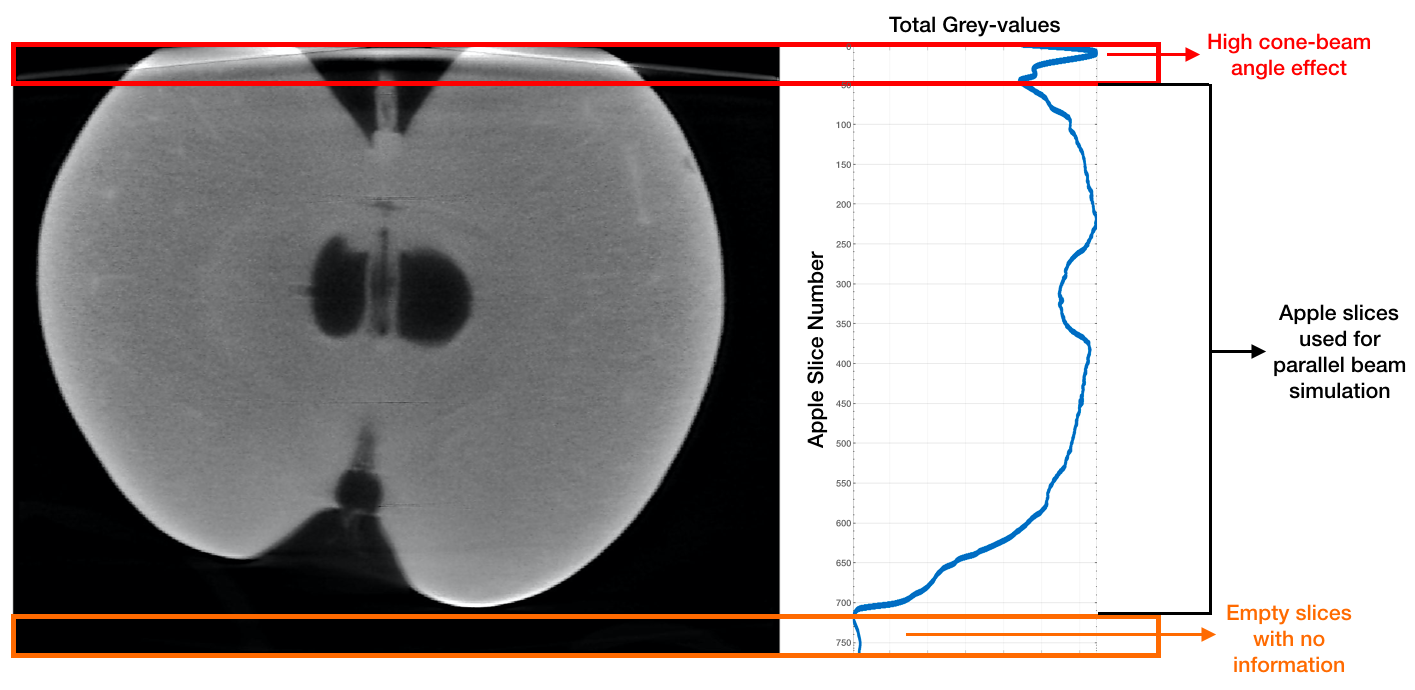}
\caption{An example vertical volume slice is given on the left with high cone beam angle artefacts visible on top of the reconstruction. The graph showing the total of grey-values per slice is given on the right. The red and orange bars highlight the slices that were discarded (50 slices from top and bottom). The corresponding graph of grey-values for the first 50 slices also show the spike of high cone-beam artefact on top, and the bottom 50 the lack of any apple information. The middle 668 slices were used for parallel beam simulations (Dataset A) and those with added Gaussian noise (Dataset B).}  
\label{fig:applerecon-threshold}
\end{figure}

Each apple slice was saved as an individual image, meaning we had $94\times768 = 72192$ reconstructed slices. However, not all slices carried relevant information: some slices often showed the remains of sepal or stamens of the apple, or were simply air. This meant that we had to cut out a portion at the bottom of the reconstructed volume in order to reduce the data to contain only relevant apple information. Because the apples were of the same type (and thus of similar shape and sizes), we decided to determine which slices to discard using the first apple. Each horizontal slice with all pixels close to zero (with zero denoting "air") were treated as empty slices. This was roughly equal to 50 slices at the bottom of a vertical apple slice. An illustration of this is given in Figure~\ref{fig:applerecon-threshold} where the empty slices and their corresponding total grey-values are indicated with an orange rectangular box.

We have also removed the top 50 slices as an attempt to exclude any high cone-beam artefact, which stemmed from the sample nearing the border of the cone of X-ray beams. Apples that were bigger than the average apple showed more pronounced high cone-beam artefacts as they were harder to fit into the field-of-view (without having to change the acquisition geometry). Therefore discarding the very top slices also meant the exclusion of such effect. For the first apple, the slices affected by the high cone-beam artefact were determined by the narrow spike in the total grey-value distribution. This is indicated in Figure~\ref{fig:applerecon-threshold} with a red arrow, and the affected slices (which are later discarded) with a red box. 

\subsection*{Parallel beam projection data}

\begin{figure}[ht]
\centering
\includegraphics[scale=0.35]{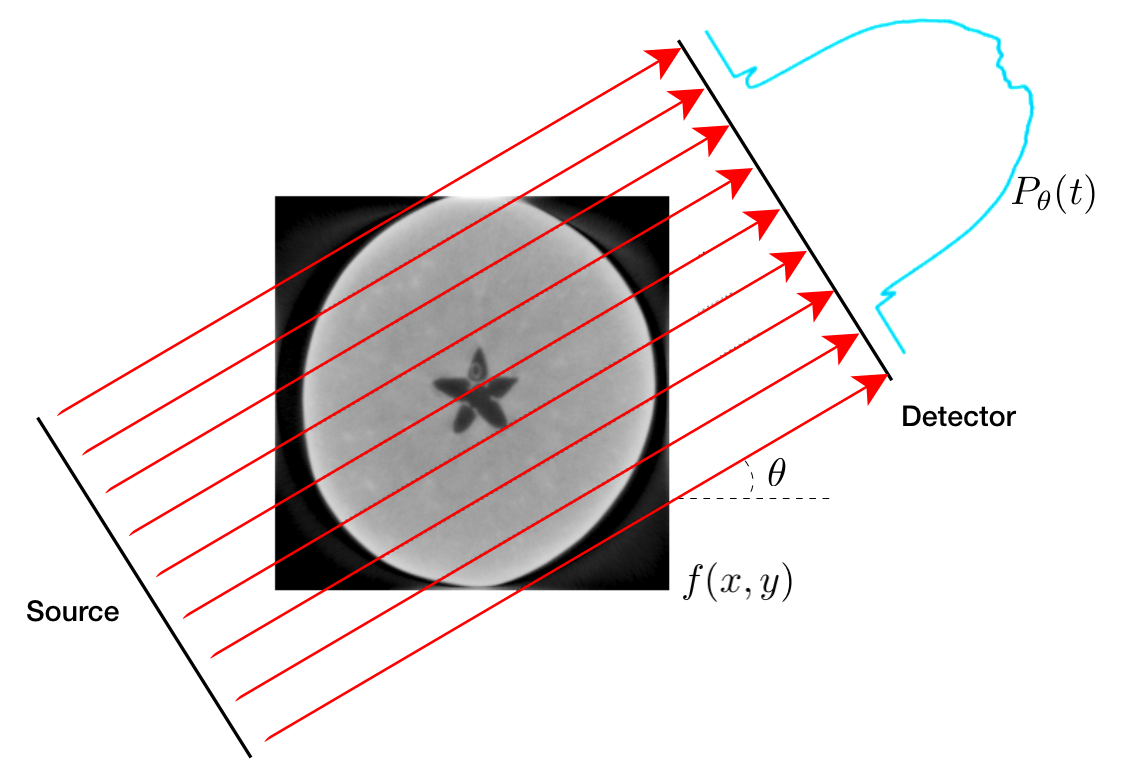}
\caption{Parallel beam projection geometry used to generate Datasets A and B}
\label{fig:par_geom}
\end{figure}

We simulated 1D parallel beam projections of each apple slice using the 2D Radon transform 
\begin{equation}
    P_\theta(t) = \int_{-\infty}^{\infty} \int_{-\infty}^{\infty} \mu(x,y) \delta(x\cos\theta + y\sin\theta -t) dx dy,
\end{equation}
where $\mu(x,y)$ is the attenuation coefficient at the voxel grid point $(x,y)$, $\theta$ is the projection angle and $P_\theta(t)$ is the 1D projection at angle $\theta$. In practice, one has to discretise the Radon transform to apply it as a linear operator. Several discretised versions of the Radon transform are available in open-source software packages, such as the ASTRA toolbox \cite{ASTRA} and \textsc{scikit-image} \cite{van2014scikit}. We used the Operator Discretisation Library (ODL) \cite{adler2017odl} to define our projection geometry (Figure \ref{fig:par_geom}) and generated the data using the CUDA projection kernel from the ASTRA toolbox as backend. 

Using the same discretisation for simulated data generation and reconstruction amounts to committing an "inverse crime". This is because data that is generated in this way lives in the range of the forward operator, and consequently, the (noiseless) reconstruction problem is not ill-posed \cite{kaipio2007statistical}. In order to avoid this, we upscaled all ground truth slices with bilinear interpolation using \textsc{scikit-image}. These upscaled images were used to generate projection data. 

For each apple slice, projection data for 50 angles uniformly spaced in $[0, \pi)$ were generated. These are henceforth referred to as Dataset A. Dataset B was generated by corrupting Dataset A with 5\% Gaussian noise, which has been used in the literature to model electronic and thermal noise in the detector during data acquisition \cite{oliver1965thermal}. For each of the data in Dataset A, a zero-mean Gaussian distribution with a standard deviation equal to \num{5}\% of the noiseless data mean was generated in the projection space using \texttt{odl.phantom.noise.white\_noise}. This distribution was then pointwise added to the noiseless data to generate Dataset B.

\subsection*{Scattering}

As an additional challenge, we imitated an effect of X-ray photon scattering for the noiseless data (Dataset A). An accurate simulation can be done using Monte-Carlo methods. However this solution requires significant computational time to reduce relative error due to Poisson noise. Instead we have applied an approximate filter that does not precisely predict the physical effect. The filter is based on the scattering kernel superposition method and uses a simplified model from \cite{bhatia2016scattering}. The scattering signal is calculated as a sum of contributions from pencil beams passing through the object. The scattering kernel is not constant and depends on the thickness of the object along a pencil beam trajectory. Scattering intensity in a pixel $u'$ from the pencil beam corresponding to the pixel $u$ is
\begin{align}
    S(u') = T(u)G(u, u'),
\end{align}
\noindent where $T(u)$ is a forward scattering intensity, and $G(u, u')$ is defined by the angular distribution. This equation was written in a simple form
\begin{align}
    S(u') = A(u) \exp\left[-\frac{(u-u')^2}{2\sigma_1 (u)^2}\right]
          + B(u) \exp\left[-\frac{(u-u')^2}{2\sigma_2 (u)^2}\right],
          \label{eq:scattering}
\end{align}
\noindent where $|u-u'|$ is a distance between pixels, and scattering is approximated as a combination of Gaussian blur with standard deviations $\sigma_1$ and $\sigma_2$. Gaussian blur parameter depends on the X-ray absorption in the pixel $u$. 

To sample functions $A(u)$, $B(u)$, $\sigma_1(u)$ and $\sigma_2(u)$, a Monte Carlo simulation based on Geant4 toolkit\cite{agostinelli2003geant4} is performed for different thicknesses of the material. In the simulation a pencil beam with an energy spectrum corresponding to the 45kV tube passes through a cylinder made of water. Thickness of the cylinder can be varied to measure different values of intensity functions $A(u)$ and $B(u)$. A distribution of photons around the cylinder is measured to approximate values of standard deviations $\sigma_1(u)$ and $\sigma_2(u)$.

\begin{figure}[ht]
\centering
\includegraphics[scale=0.35]{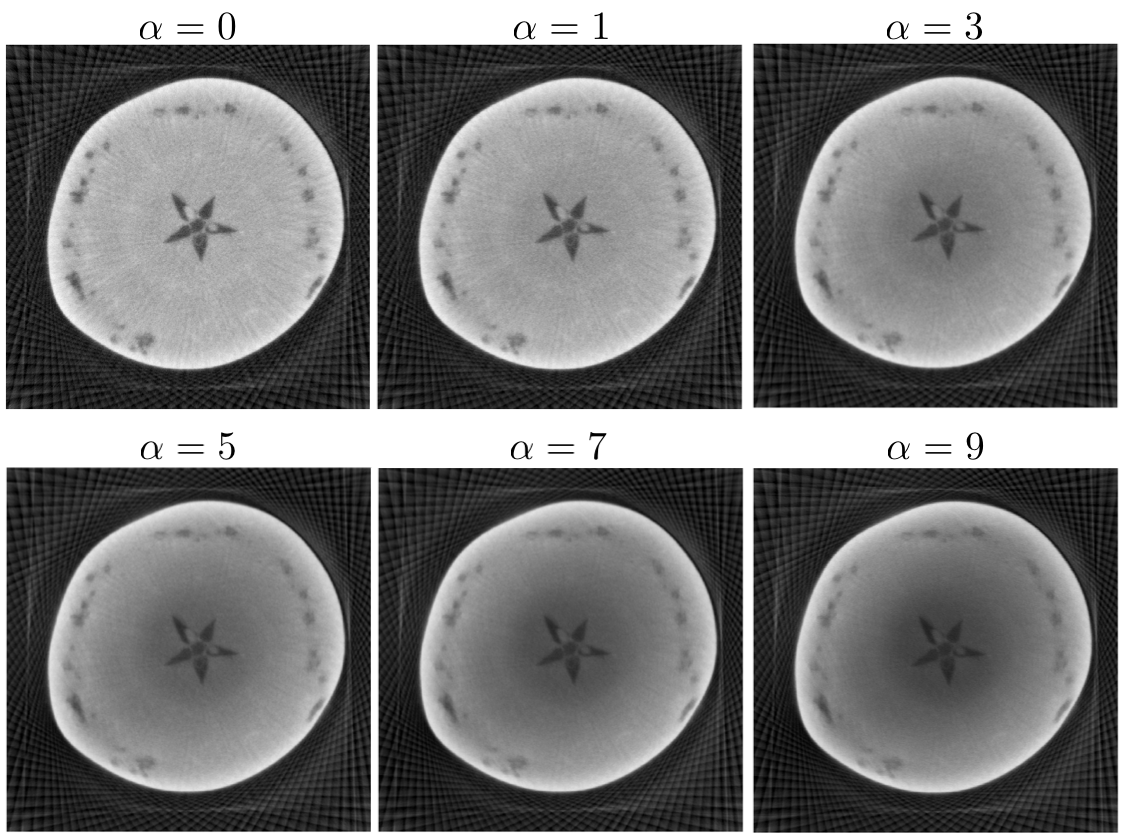}
\caption{The effect of scattering on an apple slice as the scattering intensity scaling factor, $\alpha$ is increased. Here, $\alpha = 0$ signifies no scattering, and $\alpha = 1$ the scattering of water with no additional intensity.} 
\label{fig:scatter_alpha}
\end{figure}

The approximate scattering effect was applied to the subset of slices for every apples in the dataset according to Equation \ref{eq:scattering}. We note that this filter is not expected to accurately represent the distribution of the physical effect since the scattering properties of the flesh of an apple differ from those of water. The main goal of this post-processing is to create an additional challenge for reconstruction algorithms. The functions $A(u)$ and $B(u)$ were multiplied by a scattering intensity scaling factor, $\alpha$, to enhance the effect. Figure~\ref{fig:scatter_alpha} shows the comparison of reconstructed slices with different values of $\alpha$ with $\alpha = 1$ corresponding to the scattering of water. Larger values of $\alpha$ naturally lead to stronger artefacts and noisier reconstructions. 

Slices with the approximate scattering noise are henceforth referred to as Dataset C. We note that the scattering filter was applied to Dataset A, and therefore, Dataset C contains only scattering and no other types of noise. We chose $\alpha = 5$ as the scattering intensity scaling factor for Dataset C. This effect was applied to a selection of apple slices to reduce the computation time. These slices are 195--214, 365--404, 525--544 (80 slices in total).  

\subsection*{Eliminating the label bias}

For the purpose of developing or testing learning-based reconstruction algorithms, the Apple CT datasets will have to be split into training and test subsets. In this paper, we consider a rough split of 20\% -- 80\% for the test and training subsets, respectively. To prevent any selection bias towards a specific apple, it is essential that slices from the same apple did not appear in both subsets simultaneously. Therefore the split was performed on an "apple level". We prepared the data split such that 20 apples were selected for the test subset, and the remaining 74 for the training. 

The scanned apples contain four types of defects, namely bitterpit, holes, browning and rot. We note that these are not the only defects that appear in this selection of apples, but they are the most common types found in the unhealthy parts of the samples. Therefore we primarily focus on these four defects; the smaller or more uncommon defects are counted as healthy parts of the apples.

Earlier we mentioned a "\textit{rough split of 20\% -- 80\%}" while going for a 20 apples -- 74 apples split. In truth, the main criterion of a good split is an equal percentage of the defects within the test or training subsets and not in the numbers of the apples. If one defect has a higher representation in either of the subset, the selection bias might affect the final results. Therefore, we need an equal representation of each defect in both subsets. Our goal is to find such a sequence of 20 apples that contain 20\% of each defect for the test subset, meaning the remaining 74 apples would have 80\% of each defect for the training subset.  

Along with the dataset, we include two apple label files that contain the number of pixels assigned for each defect, for all apples. These files were obtained using a label file (on slice level) provided to us by our industrial partners.  The slice-based label information was then summed per apple in order to be able to calculate defect percentages for each apple. We use this information to consider two methods of splitting the dataset into test or training subsets. 

We note here that the target defect percentage and the number of apples in a subset are independent of one another, meaning any target defect percentage with any number of apples can be achieved. However, as the percentage and the apple numbers grow farther from one another, the problem of finding an optimal sequence becomes unstable with either method explained below. With the code included in this submission, readers can perform a split of any size for any defect percentage suitable for various levels of difficulty when assessing their machine learning based reconstruction methods. 

\subsubsection*{Empirical data split}

\begin{figure}[ht]
\centering
\includegraphics[scale=0.4]{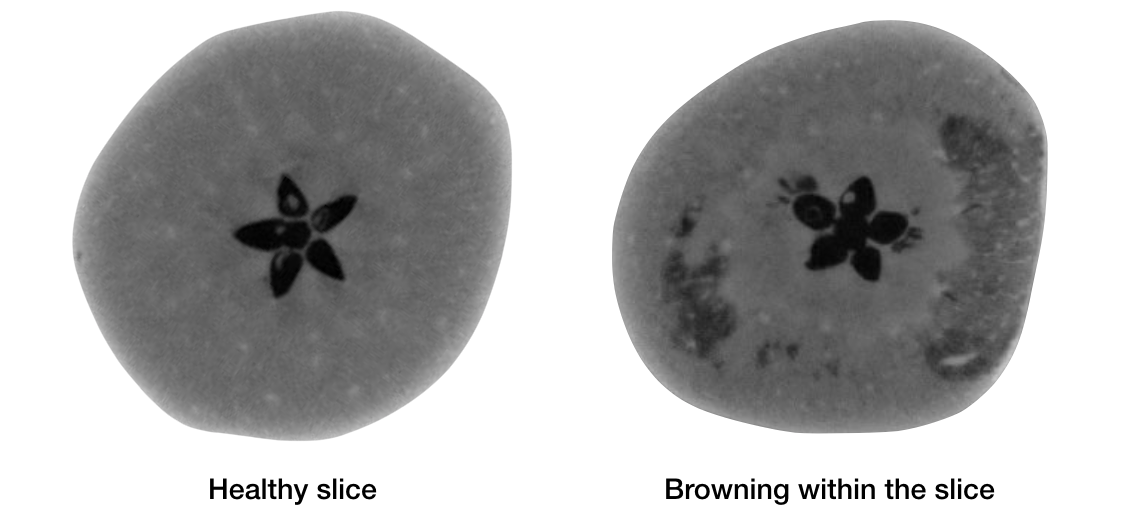}
\caption{A horizontal cross-section of a healthy apple is shown on the left, and another cross-section with browning defects of the same apple shown on the right. The red arrows point to larger browning defects across the slice.}
\label{fig:browning} 
\end{figure}

We first consider a simple heuristic algorithm, in which a random sequence of apples is tested until the target defect percentage for a selected defect is obtained while constraints are still satisfied. In our case, the defect we concentrated on was "browning" as this was the most prominent of the four defects. Figure~\ref{fig:browning} shows an example of a reconstructed slice of a healthy apple and a slice from an unhealthy apple with browning defect. Our initial aim was to reach the 20\% threshold for the browning percentage while the number of apples in sequence was exactly 20. At such a time these two conditions are satisfied, the tested sequence is saved as "successful". We used the function \texttt{randperm} in MATLAB\cite{MATLAB:2020,MATLAB:randperm} (equivalent to \texttt{numpy.random.permutation} for Python\cite{Numpy:randperm}), which takes a sequence of integers (in this case apple indices from 1 to 94) and performs a random permutation without any repeats. We performed this permutation 10000 times and recorded all "successful" sequences of apples. It is interesting to note that out of 10000 permutations, 545 were recorded as "successful". In fact, through testing, we noticed that for any given number of permutations, roughly 5.4\% of the sample size would be classified as a successful sequence. For the interested reader, we include a table of the empirical splitting results for sample size 1e$\gamma$, for $\gamma=2,\ldots,8$ in Table~\ref{tab:split_runs}. For $\gamma = 1$ and $\gamma \geq 9$ are excluded: for the former, chances of a successful sequence is very low (often not found), and for the latter, the memory required exceeds the current limit. The table includes the number of successful sequences and the CPU time it took to complete the samples. 

\begin{figure}[ht]
\centering
\includegraphics[scale=0.35]{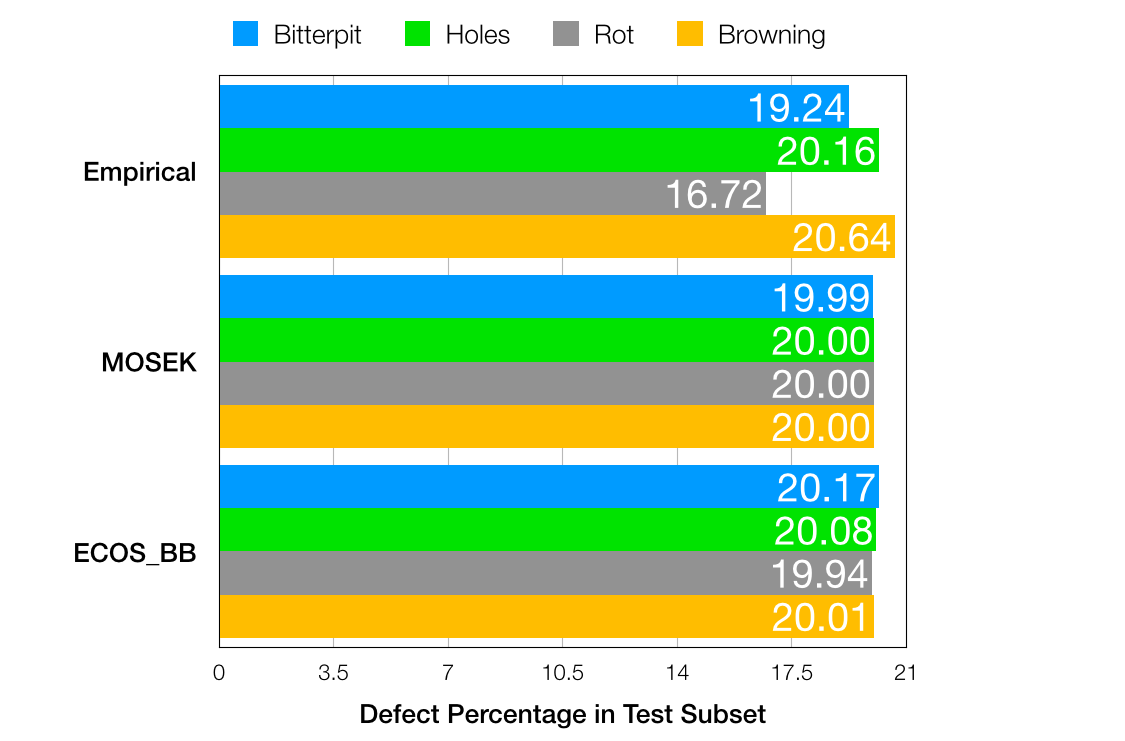}
\caption{A bar chart that shows the defect percentage distribution for all defects within the Test subset. The percentage distribution is determined using the resulting data split sequence obtained via the simple heuristic method (referred to as "Empirical") and mixed-integer quadratic programming solvers MOSEK via MATLAB ("MOSEK") and ECOS\_BB via Python, ("ECOS\_BB").}
\label{fig:full_defects_barchart}
\end{figure}

We sorted the difference of the defect percentages for each successful sequence in ascending order, and proceeded with the top result (i.e. the sequence with the least difference in defect percentages was selected). This sequence is given in the first column of Table~\ref{tab:full_datasplit_results}, with the corresponding apple numbers in the adjacent column. The defect percentage results are given in the stacked bar chart in Figure~\ref{fig:full_defects_barchart}.

\begin{figure}[ht]
\centering
\includegraphics[scale=0.35]{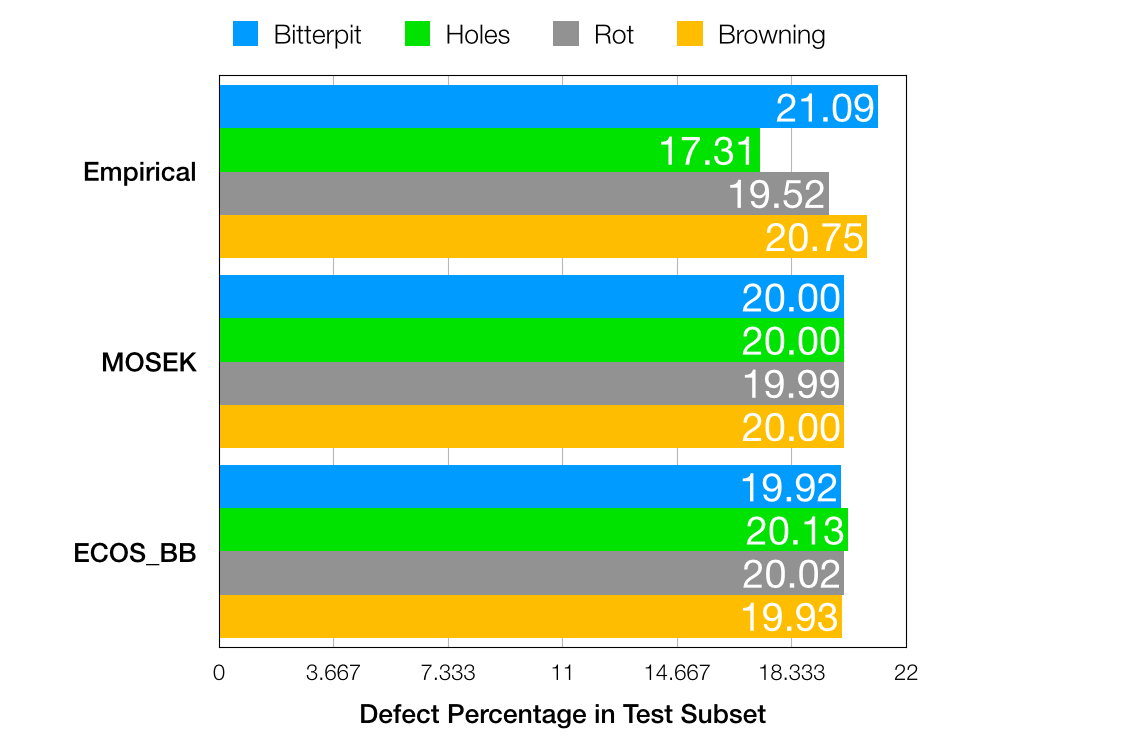}
\caption{A bar chart that shows the defect percentage distribution for all defects within the Test subset. The percentage distribution is determined using the resulting data split sequence obtained via the simple heuristic method (referred to as "Empirical") and mixed-integer quadratic programming solvers MOSEK via MATLAB ("MOSEK") and ECOS\_BB via Python, ("ECOS\_BB").}
\label{fig:partial_defects_barchart}
\end{figure}

We note that this is an empirical method of achieving a split with defect bias eliminated but certainly not the only one. We also note that the sequences for Dataset A, B and for Dataset C are different since the number of slices differ. The results of the split for Dataset C are given in Table~\ref{tab:partial_datasplit_results} with the corresponding defect percentage spread in Figure~\ref{fig:partial_defects_barchart}. 

\subsubsection*{Data split through integer programming}

The above method, introduced as "empirical", can be seen as an example of a greedy algorithm however one can also consider the data splitting as a \textit{0-1 knapsack problem}. The knapsack problem is a combinatorial optimization problem where we select items from a given set where each item has a certain weight and value. The aim is to maximize the total value while staying below a given weight threshold. The 1-0 knapsack problem is where the items are integers and are simply labelled as "in" (1) or "out" (0). The problem can be NP-hard, yet still be efficiently solved\cite{nphard}. 

Within the context of the apple data split into test and training subsets, we can formulate the problem as the following. 
\begin{align}\label{eq:split_miqp}
\begin{split}
    \min \quad & \|\sum_i c_{i,j}x_i-b_j\|^2_2, \quad i = 1,\ldots,94,\\
\textrm{s.t.} \quad & \sum_i{w_{i}x_i} = N, \textrm{ and}\\
  &x_i \in \{0,1\},    
\end{split}
\end{align}
where $b_j$ is the target percentage of each defect $j$ in the subset; $c_{i,j} \in \mathbb{R}^{94\times 4}$ is the matrix containing defect percentage information for each scanned apple, $w_i$ a vector of ones, $N$ the constraint on the number of apples in the subset and $\|\cdot\|_2$ denotes the (Euclidean) 2-norm. The array $x_i$ is the binary-valued vector of apples, where the value 1 indicates an apple to include in the subset; 0 the apple to exclude, and the index $i$ of the array denotes the specific apple corresponding to the row $i$ of $c_{i,j}$. We set our problem to obtain a sequence for the test subset, meaning we pick $b_j = 0.20, \, j = 1,\ldots,4$, for the target defect percentages, and $N = 20$ for the number of apples in the sequence. A clear advantage of this method over the empirical algorithm is that it allows for the optimisation for all defects simultaneously. 

We used CVX\cite{cvx,cvx_gb08}, a toolbox used for convex optimisation for MATLAB and Python, in conjunction with the MOSEK\cite{mosek,cvx_mosek} solver in MATLAB, and ECOS\_BB\cite{cvxpy,cvxpy_agrawal,cvxpy_ecosbb} in Python. The former requires an academic license that takes minutes to obtain, and latter being open-source. The precision for either solver was set to default, which is $\epsilon = \num{2.22e-8}$\cite{cvx_precision}. The resulting sequences for Dataset A, B and separately for Dataset C are given in Tables~\ref{tab:full_datasplit_results},\ref{tab:partial_datasplit_results}, and their corresponding defect percentage distribution in Figure~\ref{fig:full_defects_barchart},\ref{fig:partial_defects_barchart}. For more information on these specific solvers, we refer the readers to \cite{mosek:solvers,coneref_1,coneref_2}. We also detail how these solvers are employed in Usage Notes and the relevant information for the scripts in Code Availability. 

\section*{Data Records}

Dataset are made available via Zenodo\cite{appledataset}, along with two label files containing pixel numbers for each defect for all apples. Each dataset is uploaded as a compressed folder, which can be downloaded individually. Full data information can be found in Table~\ref{tab:zenodo_upload}.

\section*{Technical Validation}

The availability of large, publicly available datasets -- comprising both the raw tomographic data and corresponding ground truth image reconstructions -- is a key enabling factor for the development of data-driven learning-based reconstruction algorithms.
Such algorithms are currently being developed for various applications of tomographic imaging, in which the acquired data are limited in one or more ways, such as low-dose~\cite{chen2017low}, sparse-angle~\cite{hamalainen2013sparse}, or limited-view~\cite{ding2019joint} CT. 
In this section, we will demonstrate the feasibility of using our apple CT datasets to simulate one of these limited-data CT set-ups, namely limited-view angular sampling. Limited-view image artefacts are very common in in high-throughput~\cite{yang2011high} or conveyor belt~\cite{janssens2015neural} CT, in which full angular range data cannot be obtained due to limitations in acquisition time, apparatus set-up or hardware configurations.

\begin{figure}[ht]
\centering
\includegraphics[scale=0.425]{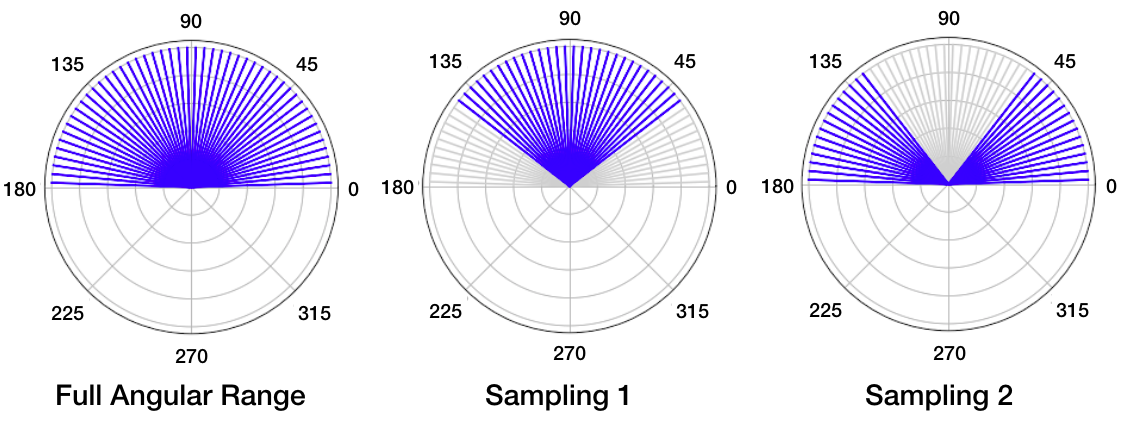}
\caption{Different angular samplings of generated parallel beam data: (\textit{left to right}) full angular range, missing wedge sampling 1 and missing wedge sampling 2. Blue lines indicate the projection angles that were selected for the limited-view data setup.}
\label{fig:mw_configs}
\end{figure}

We use our apple CT dataset to simulate two different limited-view settings. In both settings, we have data along \num{30} projection angles and a total missing wedge of \ang{75.6}. In the first setting (henceforth Sampling 1), we consider projection data between \ang{37.8} and \ang{142.2}. In the second (Sampling 2), we consider data between \ang{1.8} and \ang{52.2} as well as between \ang{127.8} and \ang{178.2}. These two configurations are shown in Figure \ref{fig:mw_configs} where the blue lines indicate an acquired view, and grey, missing.

\begin{figure}[ht]
\centering
\includegraphics[scale=0.425]{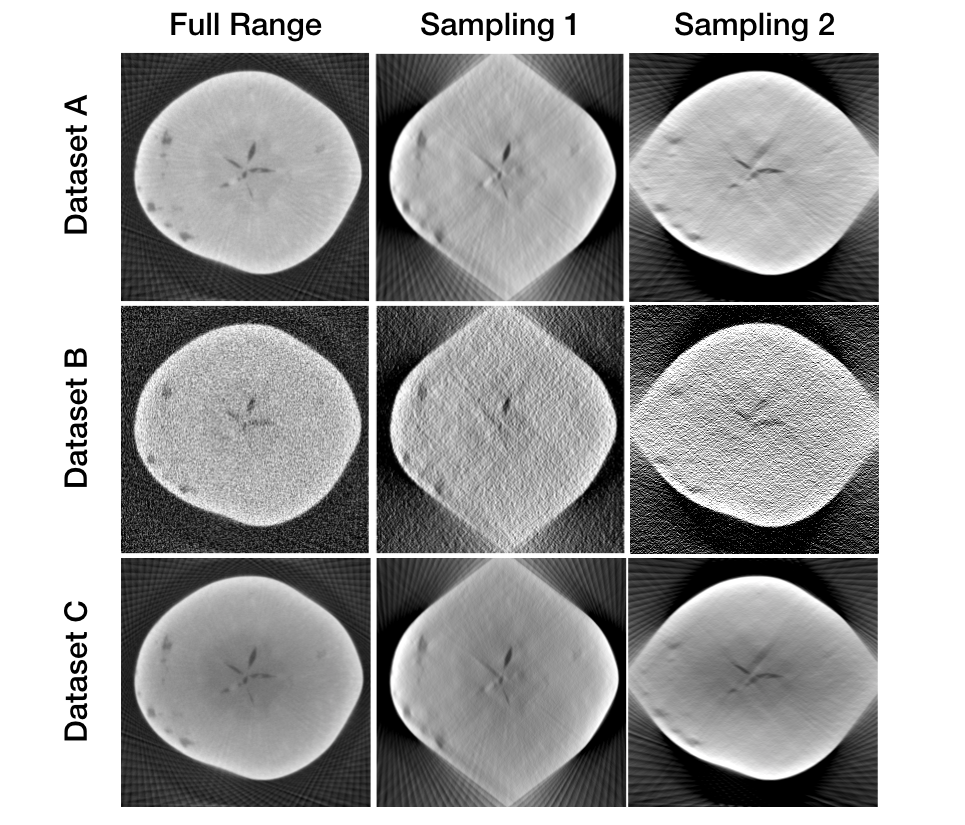}
\caption{Baseline filtered back-projection (FBP) reconstructions for full angular range and the two limited-view data setup. The grey-values of the Dataset B and C reconstructions are set to the same range of Dataset A.} 
\label{fig:baseline-recons}
\end{figure}

We computed filtered back-projection (FBP) reconstructions for Datasets A, B and C with data from the full angular range, as well as the two limited-view settings (Figure \ref{fig:baseline-recons}).
For this, we defined a reconstruction space and projection geometry using ODL, and used these to define a discretised ray transform and an FBP operator:\\ 
\indent \texttt{reco\_space = odl.uniform\_discr(min\_pt, max\_pt, reco\_shape)
\\
\indent reco\_geometry = odl.tomo.parallel\_beam\_geometry(reco\_space, num\_proj\_angles)
\\
\indent reco\_ray\_trafo = odl.tomo.RayTransform(reco\_space, reco\_geometry)
\\
\indent fbp\_op = odl.tomo.analytic.fbp\_op(reco\_ray\_trafo).
}\\
The reconstruction could then be obtained by applying the FBP operator to data:
\\
\indent \texttt{reco = fbp\_op(data)}.

Reconstructions of noisy data using FBP were poor and those of limited-view data had severe artefacts.
In the following, we show that a convolutional neural network (CNN) can be trained to improve the image quality of these FBP reconstructions by means of post-processing.
A mixed-scale dense convolutional neural network (MS-D network) recently proposed by Pelt \textit{et al.}~\cite{pelt2018improving} was subsequently trained to correct the limited-view artefacts. We used a standalone MS-D network implementation for Python with a width of 1 and a depth of 100 layers. The MS-D networks for all datasets were trained for around 90-100 epochs, and the parameters with the best validation loss were used for artefact removal. Networks for Dataset A were trained for 81 epochs (Sampling 1) and 67 epochs (Sampling 2); for Dataset B, 87 epochs (Sampling 1) and 83 epochs (Sampling 2), and for Dataset C, 116 epochs (Sampling 1) and 109 epochs (Sampling 2). Validation subset size is 1000 images for Datasets A and B, and 250 for Dataset C.
 
\begin{figure}[ht]
\centering
\includegraphics[scale=0.3]{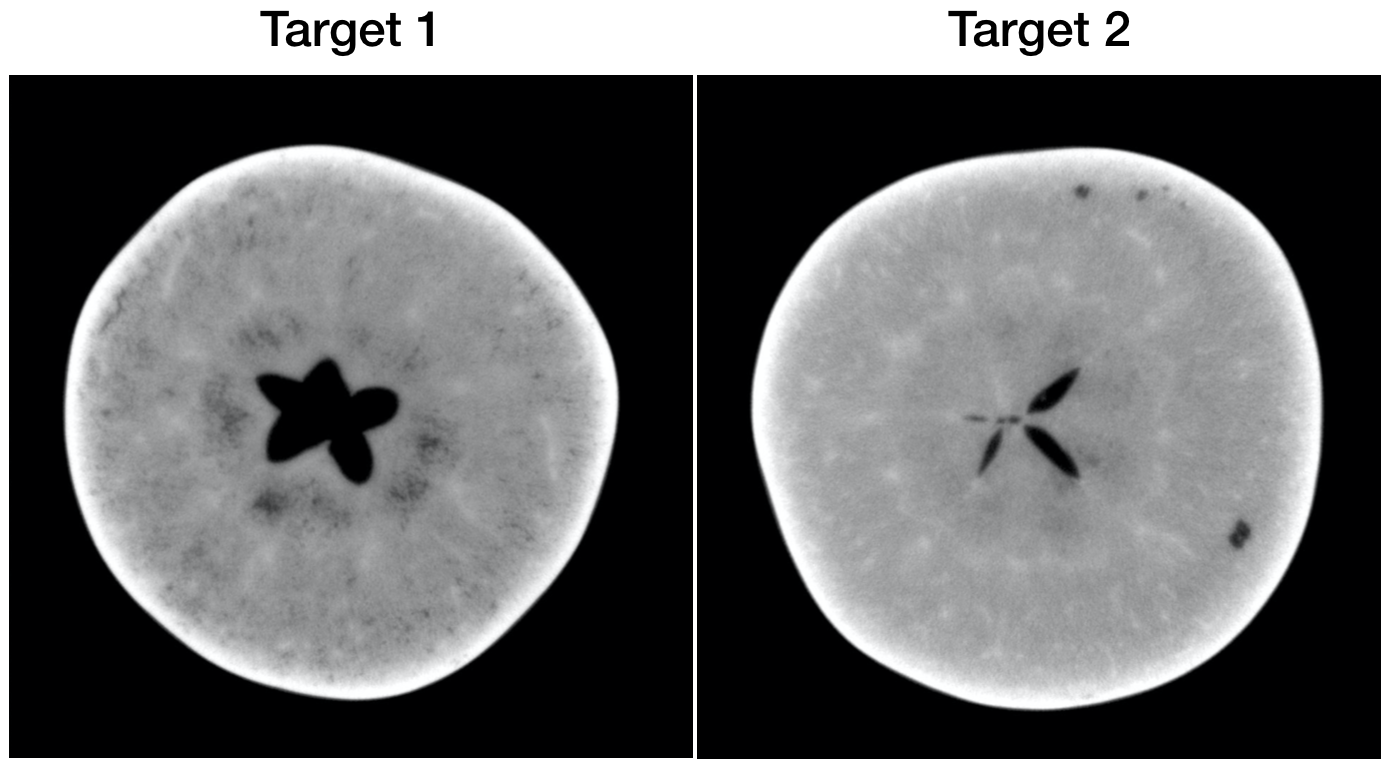}
\caption{The target apple slices for the MS-D network.}
\label{fig:msd-targetslices}
\end{figure}

\begin{figure}[ht]
\centering
\includegraphics[scale=0.425]{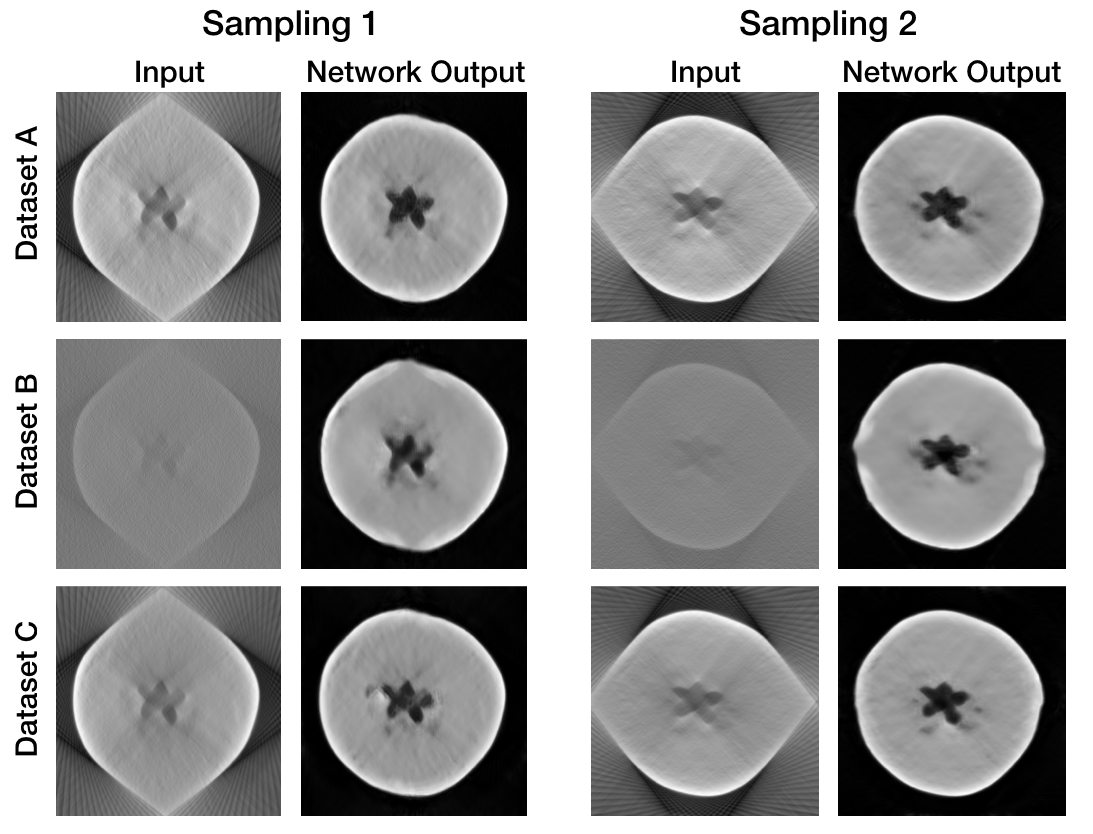}
\caption{MS-D post-processing results for the Target 1 shown in Figure~\ref{fig:msd-targetslices}. The FBP reconstructions used as inputs are given for both missing wedge angular samplings for each data, and the resulting output by the network. The grey-values of the Dataset B and C reconstructions are set to the same range of Dataset A.}
\label{fig:msd-postprocessing-target1} 
\end{figure}

\begin{figure}[ht]
\centering
\includegraphics[scale=0.425]{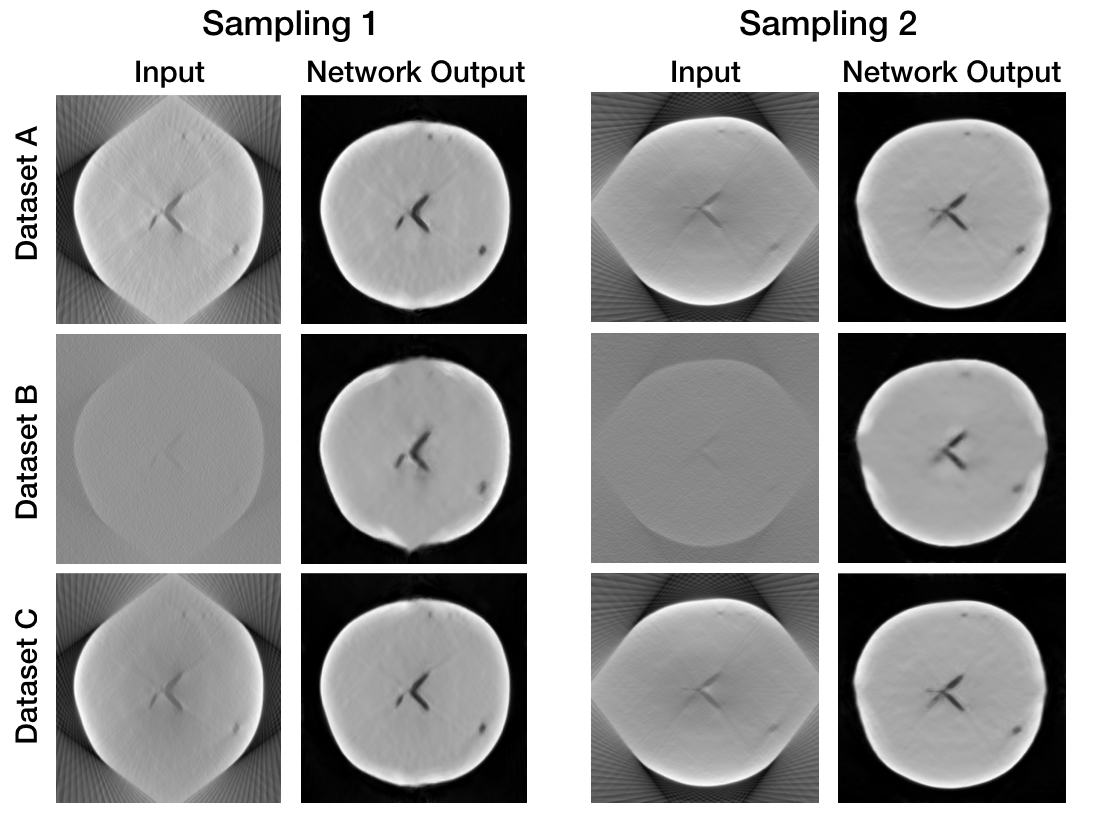}
\caption{MS-D post-processing results for the Target 2 shown in Figure~\ref{fig:msd-targetslices}. The FBP reconstructions used as inputs are given for both missing wedge angular samplings for each data, and the resulting output by the network. The grey-values of the Dataset B and C reconstructions are set to the same range of Dataset A.}
\label{fig:msd-postprocessing-target2} 
\end{figure}

In order to demonstrate the results of the training, we include the FBP input and the output of post-processing for two random apple slices. The ground truth reconstructions of these slices are given in Figure~\ref{fig:msd-targetslices}. The FBP reconstructions for Datasets A-C (the input), and the corresponding MS-D post-processing results (the network output) for both limited-view angular sampling are shown in Figures~\ref{fig:msd-postprocessing-target1} and \ref{fig:msd-postprocessing-target2}. Visually, the improvements on the input images are clear: it is evident that the network has learned the general shape of the apples, and therefore manages to compensate for the missing views well. The inner texture of the apples is not fully recovered, though the defects in either slice are recognisable by eye. In addition, we note the shape of the core of the apples, which still exhibit some influence of the missing view artefacts (stretched out length-wise for Sampling 1, width-wise for Sampling 2). This affect is particularly pronounced for Sampling 2, Dataset B in both slices. In fact, in the cases of Dataset B, the location and shape of defects are still heavily influenced by the missing view artefacts. This brings up the question how reliable MS-D network is in a practical setting with more unpredictable or amplified noise, or what else can be done to improve the input image or the network performance in order to accurately identify the size, shape and location of defects within an apple slice. 

We employed peak signal-to-noise ratio (PSNR) and structural similarity index measure (SSIM) to assess the improvements in image quality. These measures are included for both the input slices (the FBP reconstructions) and the MS-D network output. The results given in Table~\ref{tab:msd-results} are the mean and standard deviation of PSNR and SSIM results based on 15360 slices for Datasets A and B, and 1600 slices for Dataset C. We must note that these measures are not indicators of how accurate a physical result can be extracted from these reconstructions, but rather how similar an image is (in terms of grey-value distribution and structural similarity) to the ground truth image. The PSNR and SSIM results validate the improvements we saw in the apple slices: the results improve drastically upon learned post-processing, with the output reconstructions for Dataset A and C approaching the image quality of the ground truth, and those for Dataset B following closely behind.

\section*{Usage Notes}

The results presented in this manuscript are obtained using scripts written in MATLAB and Python programming languages. The scripts are tested with MATLAB 2017a,b to 2020a, and Python 3.6.10 to 3.7.7, and are distributed under the MIT license. We make use of the following toolboxes: ASTRA\cite{ASTRA}, ODL\cite{adler2017odl}, CVX\cite{cvx,cvxpy} (MOSEK\cite{cvx_mosek} and ECOS\_BB\cite{cvxpy_ecosbb}) and Beads Dataset Project\cite{beadscodes,beadsmanual}. 

ODL and the ASTRA toolbox can be installed in an appropriate Python environment using \texttt{conda} by running \texttt{conda install -c odlgroup odl} and \texttt{conda install -c astra-toolbox astra-toolbox}, respectively. To use the latest development version of the ASTRA toolbox, run \texttt{conda install -c astra-toolbox/label/dev astra-toolbox}. Detailed installation instructions \footnote{https://github.com/astra-toolbox/astra-toolbox}\footnote{https://github.com/odlgroup/odl} are available in the respective GitHub repositories. 

CVX download and installation notes for MATLAB are found in \footnote{http://cvxr.com/cvx/download/} \footnote{http://cvxr.com/cvx/doc/install.html}. The steps can be summarised as the following: (1) Download and unzip the appropriate package anywhere (we suggest the home directory); (2) Install any additional licenses for solvers (in our case MOSEK, more on this later), (3) Add the full path to the unzipped location (the folder will be named "cvx"). You can add path using the \texttt{addpath} command in MATLAB. This is akin to importing a package in Python, and is required for any external toolboxes. Finally, (4) install CVX by typing \texttt{cvx\_setup} in the console (Command Window). For Python, you can install CVXPY in a suitable environment by using \texttt{pip} or \texttt{conda} \footnote{https://www.cvxpy.org/install/}, using \texttt{pip install cvxpy} or \texttt{conda install -c conda-forge cvxpy}.

MOSEK solver is downloaded via the CVX toolbox however it requires an additional license. Readers can request a Personal Academic License\footnote{https://www.mosek.com/products/academic-licenses/} using an email address belonging to any academic institute. The license named "\textit{mosek.lic}" is emailed within minutes. Create a folder "\textit{mosek}" and place the license file prior to installing CVX. In Python, the ECOS\_BB solver\footnote{https://github.com/embotech/ecos} is installed along with the default installation of CVXPY. However, it is not the default solver and users are required to specify \texttt{solver='ECOS\_BB'} in \texttt{prob.solve}\footnote{https://www.cvxpy.org/tutorial/advanced/index.html}.

\section*{Code Availability}

The latest versions of the scripts can be found on the Computational Imaging group GitHub repository\footnote{https://github.com/cicwi/applect-dataset-project}. The label files \textit{apple\_defects\_full.csv} and \textit{apple\_defects\_partial.csv} contain the total number of labelled pixels for all 8 defects per apple, as well as healthy pixels and the background. These files are required for some of the scripts, which we introduce below. For ease of readability, the scripts are categorized into the following subsections: {\it Dataset generation, bias elimination,} and {\it reconstruction}. 

\subsection*{Dataset generation}
Here, we list the scripts used to obtain all three versions of the simulated datasets.

\subsubsection*{generate\_parallel\_data.py}
This script can be used to generate parallel-beam datasets similar to Datasets A and B from ground-truth reconstructions. It contains the definition of our parallel beam geometry using ODL, with the ASTRA \texttt{cuda} projector as backend. The script requires the user to specify the directory where the ground truth reconstructions are saved as well as the directory names for storing the noiseless and noisy data. This can be done by setting the options in the function call as follows:
\\
\indent \texttt{python generate\_parallel\_data.py -s 'source\_dir' -d 'noiseless\_data\_dir' \\
\indent -n 'noisy\_data\_dir'}\\

\subsubsection*{main.py}
Scattering generation is performed using a Python script with C program utilized for scattering calculations. The C code can be compiled using the attached Makefile by simply running the following command:\\
\indent\indent \texttt{make}\\
Scattering can be calculated for a set of noiseless projections by executing the command\\
\indent\indent \texttt{python main.py -i /path/to/input/ -o /path/to/output/ -a 5.0 -b 375 -e 385}\\
which loads the projections in the folder specified by the "-i" option; multiplies the scattering effect by the intensity factor (in the case of Dataset C, this was $alpha = 5$), and saves the results in the output folder. The output folder is expected to contain two sub-folders, \textit{log\_data} and \textit{raw\_data}. Scattering is applied to a selected range of rows, which are set using the "-b" and "-e" options (by default, this is set to rows 375--385). \\

\subsection*{Bias elimination}

For the purpose of developing algorithms based on learning information from a training set or assess a result against a test or a validation set, we have to split the data in such a way that we avoid introducing any bias towards a specific apple or a defect within that apple. We described two methods in Eliminating bias: a simple heuristic method (referred to as "Empirical") and mixed-integer quadratic programming using MOSEK via MATLAB ("MOSEK") and ECOS\_BB via Python,("ECOS\_BB"). Here we list the scripts used to solve this problem. Users can obtain the sequences given in Tables~\ref{tab:full_datasplit_results},\ref{tab:partial_datasplit_results} and the defect percentages in Figures~\ref{fig:full_defects_barchart},\ref{fig:partial_defects_barchart} by following the instructions below. 

\subsubsection*{datasplit\_empirical.m and datasplit\_empirical.py}
These scripts are equivalent, and therefore require the same input. The output differs due to the element of randomisation but the problem is solved in the same way. The scripts require four input variables: Path location to the data file with defect information for each apple (required); sample size for the number of times the apple index is permuted (optional, default is 10000 samples), target percentage for the browning defect within the subset, i.e. Subset 1 (optional, default is 0.2, as in 20\%), and required number of apples in the subset (optional, default is 20 apples). Once Subset 1 is determined, the remaining apples are considered as part of Subset 2. \\
For a 30\%, 10 apples test -- 70\%, 84 apples training data split, running the following commands in MATLAB and in Python would return similar defect percentages given in Figure~\ref{fig:full_defects_barchart}. \\
\indent\indent \texttt{datasplit\_empirical(filepath, 10000, 0.3, 10)}\\
\indent\indent \texttt{python datasplit\_empirical.py -i filepath -s 10000 -p 0.3 -n 10}\\
\noindent where \textit{filepath} is the location of the defect data file "{\it apple\_defects\_full.csv}" is saved. This is valid for Dataset A and B. For Dataset C, the required input file is "{\it apple\_defects\_partial.csv}".\\
The script returns two data files (in csv format): the apple defect information for the sequence determined as Subset 1, and the remaining apples as Subset 2. Files are saved to the current folder. In addition, a text file named "\textit{empirical\_split\_results.txt}" is saved, which contains the number of successful sequences, the best sequence for Subset 1 and 2 and their defect percentages.\\
Script also prints the following messages to the console:\\
\indent\indent \textsf{Results for split percentage 0.30 are saved in empirical\_split\_results\_subset1.csv.}\\
\indent\indent \textsf{Results for split percentage 0.70 are saved in empirical\_split\_results\_subset2.csv.}

\subsubsection*{datasplit\_mosek.m} 
The script solves the problem described in Equation~\ref{eq:split_miqp} using CVX via MOSEK solver. The problem is solved by executing the command\\
\indent\indent \texttt{datasplit\_mosek(filepath, 0.4, 30)}\\
which would return a qualifying sequence for Subset 1 containing 30 apples with 40\% defect percentage distribution, and the remainder included in Subset 2. The only required input is the \textit{filepath}: the target defect percentage value and number of apples are optional, with the same default values as in \texttt{datasplit\_empirical}. The script assumes CVX is added to the current folder path, and will quit with an error otherwise. 
The script saves three files: defect data for both subsets and a text file named "\textit{mosek\_split\_results.txt}" with the solver termination details (optimal value and CPU time) as well as the defect percentage spread and the sequence for both subsets. The console output for above command is\\
\indent\indent \textsf{Results for split percentage 0.40 are saved in mosek\_split\_results\_subset1.csv.}\\
\indent\indent \textsf{Results for split percentage 0.60 are saved in mosek\_split\_results\_subset2.csv.}

\subsubsection*{datasplit\_ecos.py}
The script solves the problem described in Equation~\ref{eq:split_miqp} using CVX via ECOS\_BB solver. The problem is solved by executing the command\\
\indent\indent \texttt{python datasplit\_ecos.py -p 0.4 -n 25 -i filepath}\\
which would return a qualifying sequence for Subset 1 containing 25 apples with 40\% defect percentage distribution, and the remainder in Subset 2. The default values are the same as for the previous scripts. The console and file output are as given above, with the word "\textit{mosek}" replaced by "\textit{ecos}" in the file names.

\subsection*{Reconstruction}
Lastly, we include the scripts used to obtain the results presented in Technical Validation. 

\subsubsection*{reconstruct\_fbp.py}
This script was used to obtain the filtered back-projection (FBP) reconstructions of all datasets for full angular range as well as the two limited-view angular samplings (Figure~\ref{fig:baseline-recons}). The script requires the user to specify the data directory with the option "-d". The directories where the reconstructions will be saved must also be provided with option "-f". This can be done by running the following command\\
\indent \texttt{python reconstruct\_fbp.py -d 'data\_dir' -f 'full\_range\_reco\_dir'}
\\
\indent \texttt{-m 'sampling1\_reco\_dir' -n 'sampling2\_reco\_dir'}

\subsubsection*{train.py}
This script trains an MS-D network that corrects limited-angle artefacts, demonstrated in Technical Validation. This can be done by running the command\\
\indent\indent \texttt{python train.py -v 1000 -l 100 -i /path/to/input -t /path/to/target}\\
which starts MS-D network training with 100 layers, 1000 images in the validation subset. Options "-i" and "-t" are used to set the location of input and ground truth reconstructions, respectively. Training process runs until the script execution is stopped by the user manually. Every 1000 images the script performs a  validation step, and saves the network parameters if a smaller loss value is achieved.

\subsubsection*{test.py}
This script tests a trained MS-D network. This script should be executed in the same folder that was used for network training. It uses network parameters from the \textit{'regr\_params.h5'} file. Test can be executed by running the command\\
\indent\indent \texttt{python test.py -i /path/to/input -t /path/to/target -v}\\
where "-i" and "-t" options are used to set the location of input and ground truth reconstructions, respectively. If the "-v" option is included, the script will save network output and comparison images into the \textit{results/} folder. The script outputs average values of PSNR and SSIM for the input and the network output reconstructions.

\subsubsection*{test\_scattering.py}
This script tests an MS-D network trained on limited-view reconstruction from Dataset C. The script is similar to \textit{test.py}, but only a subset of slices is taken from the ground truth reconstructions. It can be executed by running the command\\
\indent\indent \texttt{python test\_scattering.py -i /path/to/input -t /path/to/target -v}\\

\bibliography{appleCT}

\section*{Acknowledgements} 
Authors are grateful for the assistance of our industrial partners, GREEFA, especially Anna-Katharina Trull and Floris Berendsen for providing the apple defect label files. S.B.C. would like to acknowledge the financial support of the Netherlands Organisation for Scientific Research (NWO; project number 639.073.506); P.S.G. the Marie Skłodowska-Curie Innovative Training Network MUMMERING (grant agreement no. 765604) and M.v.E. and K.J.B. the financial support by Holland High Tech through the PPP allowance for research and development in the HTSM topsector.

\section*{Author contributions statement}
Ground truth reconstructions were generated by S.B.C. Parallel beam data generation was performed by P.S.G. Scattering was performed by V.A. Bias elimination methods were developed by S.B.C, V.A. and P.S.G. Technical validation results were obtained by P.S.G, V.A. and M.v.E. All figures and tables were generated by S.B.C, P.S.G. and V.A. All authors have contributed to text either by direct input or through comments. 

\section*{Competing interests} 
The authors declare no competing interests, financial or otherwise.

\section*{Figures \& Tables}

\begin{table}[ht]
\centering
\begin{tabular}{|c|c|c|c|c|c|c|}\cline{1-7}
\multirow{2}{*}{\bf Sample Size} & \multirow{2}{*}{\bf Successful Runs} & \multirow{2}{*}{\bf Time (s)} & \multicolumn{4}{c|}{\bf Defect Percentage} \\ \cline{4-7} 
& & & \textit{Bitterpit} & \textit{Holes} & \textit{Rot} & \textit{Browning}\\
\hhline{|=|=|=|=|=|=|=|}
\num{1e2} & 4 & 0.0024 & 12.97\% & 35.04\% & 13.38\% & 20.02\% \\ \cline{1-7}
\num{1e3} & 43 & 0.0291 & 22.01\% & 23.26\% & 17.31\% & 20.23\% \\ \cline{1-7}
\num{1e4} & 545 & 0.060 & 19.24\% & 20.16\% & 16.72\% & 20.64\% \\ \cline{1-7}
\num{1e5} & 5418 & 0.4074 & 21.62\% & 19.79\% & 19.50\% & 20.10\% \\ \cline{1-7}
\num{1e6} & 54322 & 5.1885 & 20.00\% & 20.33\% & 20.86\% & 20.02\% \\ \cline{1-7}
\num{1e7} & 542670 & 410.3428 & 20.03\% & 20.23\% & 19.87\% & 20.46\% \\ \cline{1-7}
\num{1e8} & 5431077 & 5236.5134 & 20.05\% & 20.05\% & 20.15\% & 20.02\% \\ \cline{1-7}
\end{tabular}
\caption{\label{tab:split_runs}The results of the test runs for the empirical data splitting method as the sample size is increased.}
\end{table}

\begin{table}[ht]
\centering
\begin{tabular}{||c|c|c|c|c|c||}\cline{1-6}
\multicolumn{2}{||c|}{\bf Empirical} & \multicolumn{2}{c}{\bf MOSEK} & \multicolumn{2}{|c||}{\bf ECOS\_BB} \\ 
\cline{1-6} 
\textit{Sequence} & \textit{Apple No.} & \textit{Sequence} & \textit{Apple No.} & \textit{Sequence} & \textit{Apple No.}\\ 
\hhline{|=|=|=|=|=|=|}
8 & 31108 & 1 & 31101 & 6 & 31106 \\\cline{1-6}
9 & 31109 & 3 & 31103 & 12 & 31112 \\\cline{1-6}
10 & 31110 & 20 & 31120 & 15 & 31115 \\\cline{1-6}
17 & 31117 & 22 & 31122 & 21 & 31121 \\\cline{1-6}
20 & 31120 & 26 & 31204 & 22 & 31122 \\\cline{1-6}
22 & 31122 & 28 & 31206 & 28 & 31206 \\\cline{1-6}
30 & 31208 & 35 & 31213 & 35 & 31213 \\\cline{1-6}
32 & 31210 & 43 & 31221 & 42 & 31220 \\\cline{1-6}
34 & 31212 & 45 & 31301 & 46 & 31302 \\\cline{1-6}
44 & 31222 & 47 & 31303 & 47 & 31303 \\\cline{1-6}
52 & 31308 & 56 & 31312 & 49 & 31305 \\\cline{1-6}
58 & 31314 & 59 & 31315 & 52 & 31308 \\\cline{1-6}
59 & 31315 & 65 & 31321 & 53 & 31309 \\\cline{1-6}
69 & 32103 & 67 & 32101 & 54 & 31310 \\\cline{1-6}
71 & 32105 & 69 & 32103 & 67 & 32101 \\\cline{1-6}
75 & 32109 & 74 & 32108 & 72 & 32106 \\\cline{1-6}
76 & 32110 & 77 & 32111 & 77 & 32111 \\\cline{1-6}
81 & 32115 & 80 & 32114 & 87 & 32121 \\\cline{1-6}
83 & 32117 & 90 & 32202 & 90 & 32202 \\\cline{1-6}
92 & 32204 & 91 & 32203 & 94 & 32206 \\\cline{1-6}
\end{tabular}
\caption{\label{tab:full_datasplit_results}Sequences and the corresponding apple numbers for the Test subset of Dataset A and B for each of the data splitting method. The remaining apples are counted as part of the Training subset.}
\end{table}

\begin{table}[ht]
\centering
\begin{tabular}{||c|c|c|c|c|c||}\cline{1-6}
\multicolumn{2}{||c|}{\bf Empirical} & \multicolumn{2}{c}{\bf MOSEK} & \multicolumn{2}{|c||}{\bf ECOS\_BB} \\ 
\cline{1-6} 
\textit{Sequence} & \textit{Apple No.} & \textit{Sequence} & \textit{Apple No.} & \textit{Sequence} & \textit{Apple No.}\\
\hhline{|=|=|=|=|=|=|}
2 & 31102 & 6 & 31106 & 8 & 31108 \\\cline{1-6}
7 & 31107 & 12 & 31112 & 9 & 31109 \\\cline{1-6}
12 & 31112 & 17 & 31117 & 13 & 31113 \\\cline{1-6}
13 & 31113 & 26 & 31204 & 21 & 31121 \\\cline{1-6}
14 & 31114 & 27 & 31205 & 27 & 31205 \\\cline{1-6}
32 & 31210 & 35 & 31213 & 29 & 31207 \\\cline{1-6}
36 & 31214 & 39 & 31217 & 31 & 31209 \\\cline{1-6}
43 & 31221 & 40 & 31218 & 34 & 31212 \\\cline{1-6}
45 & 31301 & 46 & 31302 & 40 & 31218 \\\cline{1-6}
51 & 31307 & 50 & 31306 & 41 & 31219 \\\cline{1-6}
52 & 31308 & 54 & 31310 & 50 & 31306 \\\cline{1-6}
59 & 31315 & 57 & 31313 & 52 & 31308 \\\cline{1-6}
68 & 32102 & 62 & 31318 & 53 & 31309 \\\cline{1-6}
69 & 32103 & 65 & 31321 & 62 & 31318 \\\cline{1-6}
73 & 32107 & 76 & 32110 & 65 & 31321 \\\cline{1-6}
81 & 32115 & 77 & 32111 & 73 & 32107 \\\cline{1-6}
84 & 32118 & 79 & 32113 & 76 & 32110 \\\cline{1-6}
87 & 32121 & 83 & 32117 & 79 & 32113 \\\cline{1-6}
90 & 32202 & 85 & 32119 & 81 & 32115 \\\cline{1-6}
93 & 32205 & 87 & 32121 & 89 & 32201 \\\cline{1-6}
\end{tabular}
\caption{\label{tab:partial_datasplit_results}Sequences and the corresponding apple numbers for the Test subset of Dataset C for each of the data splitting method. The remaining apples are counted as part of the Training subset.}
\end{table}

\begin{table}[ht]
\centering
\begin{tabular}{|c|c|c|c|}\cline{1-4}
\textbf{Dataset Name} & \textbf{Number of Apple Slices} & {\bf Slice File Name} & \textbf{Size}\\\hline
Dataset A & $668\times94=62792$ & \textit{data\_appleNo\_sliceNo.tif} & 19GB \\\cline{1-4}
Dataset B & $668\times94=62792$ & \textit{data\_noisy\_appleNo\_sliceNo.tif} & 19GB\\\cline{1-4}
Dataset C & $80\times94=7520$ & \textit{data\_appleNo\_sliceNo.tif} & 2GB \\\cline{1-4}
\begin{tabular}{c}\vspace{-0.2cm}Ground Truth\\Reconstructions\end{tabular} & \begin{tabular}{c}\vspace{-0.2cm}$768\times94=72192$\\(3D Volumes)\end{tabular} & \textit{appleNo\_sliceNo.tif} & 255GB \\\cline{1-4}
\end{tabular}
\caption{\label{tab:zenodo_upload}Zenodo data upload information for each compressed folder. The datasets can be downloaded individually.}
\end{table}

\begin{table}[ht]
\centering
\begin{tabular}{|c|c|c|c|c|c|}
\hline
\multirow{2}{*}{\begin{tabular}[c]{@{}c@{}}\bf Dataset\\\bf Name\end{tabular}} & \multirow{2}{*}{\begin{tabular}[c]{@{}c@{}}\bf Sampling\\ \bf Range\end{tabular}} & \multicolumn{2}{c|}{\bf PSNR} & \multicolumn{2}{c|}{\bf SSIM}\\ \cline{3-6} 
 & & \it input  & \it output  & \it input  & \it output \\ \hline
\multirow{2}{*}{Dataset A} & Sampling 1   & 15.1 $\pm$ 5.3  & 29.4 $\pm$ 3.1 &  0.29 $\pm$ 0.09  & 0.74 $\pm$ 0.04   \\ \cline{2-6} 
 & Sampling 2  & 15.1 $\pm$ 5.3 & 29.4 $\pm$ 3.0  & 0.30 $\pm$ 0.09 & 0.75 $\pm$ 0.04 \\ \hline
\multirow{2}{*}{Dataset B} & Sampling 1   & 1.6 $\pm$ 7.9  & 27.2 $\pm$ 3.1 & 0.01 $\pm$ 0.05  & 0.69 $\pm$ 0.05 \\ \cline{2-6} 
 & Sampling 2  & 1.6 $\pm$ 7.9 & 27.5 $\pm$ 3.2 & 0.01 $\pm$ 0.05 & 0.72 $\pm$ 0.04 \\ \hline
\multirow{2}{*}{Dataset C} & Sampling 1 & -44.8 $\pm$ 0.7 & 28.8 $\pm$ 1.9 & (3.5 $\pm$ 1.2)\num{e-6} & 0.73 $\pm$ 0.02 \\ \cline{2-6} 
 & Sampling 2 & -44.8 $\pm$ 0.7 & 28.8 $\pm$ 1.9 & (3.5 $\pm$ 1.1)\num{e-6} & 0.75 $\pm$ 0.02 \\ \hline
\end{tabular}
\caption{\label{tab:msd-results}Peak signal-to-noise (PSNR) and structural similarity index measure (SSIM) for the FBP reconstruction prior to MS-D post-processing step (the input), and the MS-D network output after. The mean and standard deviation figures are based on 15360 slices for Dataset A and B, and 1600 slices for Dataset C.}
\end{table}

\end{document}